\pdfoutput=1

\documentclass[11pt]{article}

\usepackage[]{acl}

\usepackage{times}
\usepackage{latexsym}

\usepackage[T1]{fontenc}

\usepackage[utf8]{inputenc}

\usepackage{microtype}

\usepackage{blindtext}
\usepackage{times}
\usepackage{url}
\usepackage{latexsym}
\usepackage{natbib}
\usepackage{float}
\usepackage{hyperref}
\usepackage{CJKutf8,graphicx,float,wrapfig,color,latexsym,gensymb,booktabs,mathpazo,mathtools,amsmath,amssymb,url,csquotes,tipa,enumitem,tikz,tikz-qtree,tikz-dependency,rotating,hyperref,cleveref,array,titlesec,lineno,adjustbox,multirow,supertabular,tabularx,ragged2e,booktabs,caption} 
\usepackage{flafter} 
\usepackage{tabularx}

\title{Quantifying Discourse Support for Omitted Pronouns}

\author{Shulin Zhang$^1$, Jixing Li$^2$, John Hale$^1$ \\
         $^1$University of Georgia, US \\
         $^2$City University of Hongkong, China \\
         \texttt{shulin.zhang@uga.edu}\\
         \texttt{jixingli@cityu.edu.hk}\\
         \texttt{jthale@uga.edu}\\}

\begin{document}
\begin{CJK}{UTF8}{gbsn} 

\maketitle
\begin{abstract}
\textit{Pro}-drop is commonly seen in many languages, but its discourse motivations have not been well characterized. Inspired by the topic chain theory in Chinese, this study shows how character-verb usage continuity distinguishes dropped pronouns from overt references to story characters. We model the choice to drop \textit{vs.} not drop as a function of character-verb continuity. The results show that omitted subjects have higher character history-current verb continuity salience than non-omitted subjects. This is consistent with the idea that discourse coherence with a particular topic, such as a story character, indeed facilitates the omission of pronouns in languages and contexts where they are optional.
\end{abstract}

\section{Introduction}
\begin{figure*}
    \centering
    \includegraphics[scale = 0.27]{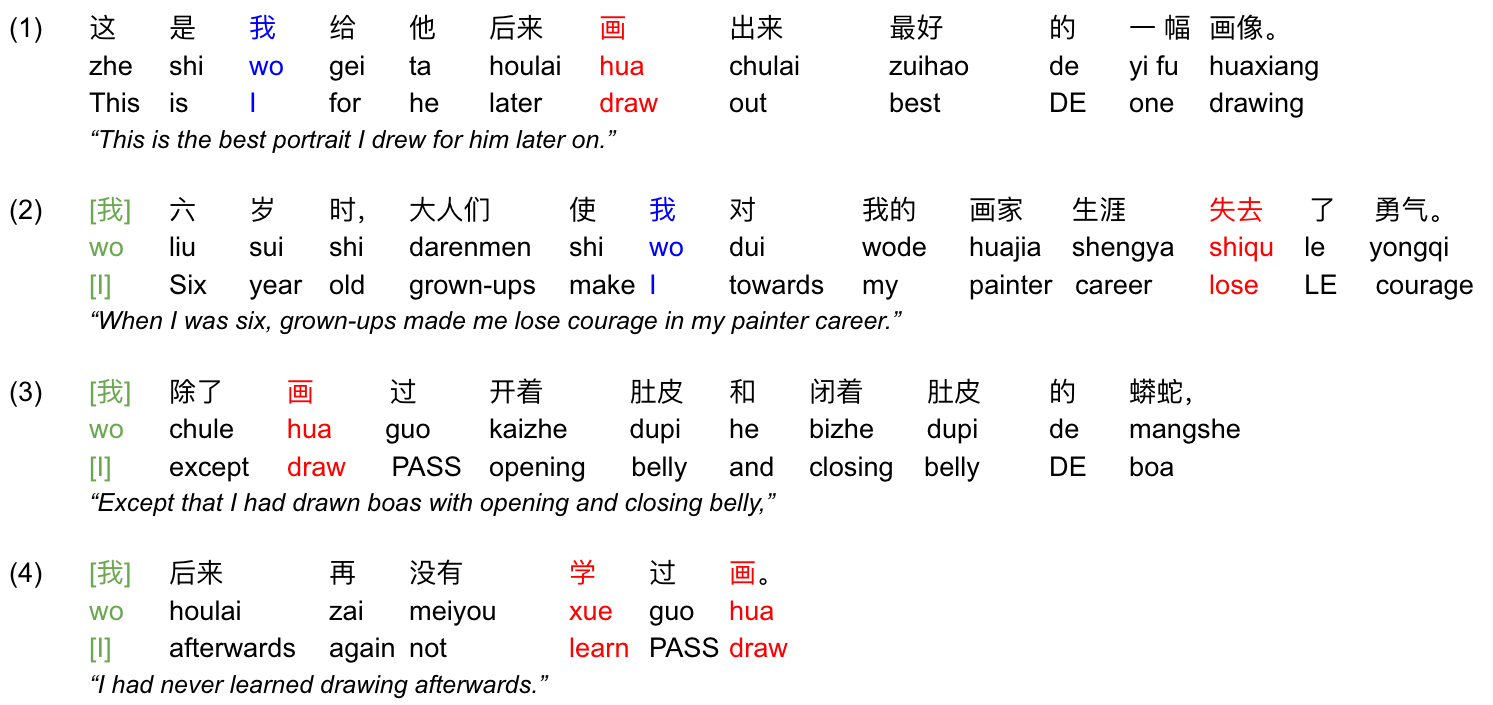}
    \caption{Example of Chinese omitted pronouns in a topic chain. Omitted pronouns, shown here in green with square brackets are not actually spoken. However, their intended reference is unambiguous for native speakers. Predicates are shown in red, and the overtly expressed entities are shown in blue. Unlike in Romance languages, there is no morphological change on verbs to mark the gender or number of omitted elements in Chinese.}
    \label{Fig:example}
\end{figure*}

\textit{Pro}-drop is a phenomenon that pronouns can be omitted when they are inferable. It is common across the world's languages, and Mandarin Chinese is one of them (See examples (3) and (4) in Figure \ref{Fig:example}). Omitted pronouns in these languages, also called zero pronouns, are increasingly important in computational linguistics (\textit{e.g.} \citealp{chen2021tackling, iida2006exploiting, iida2015intra, kong2019chinese}). This paper formalizes the notion of Topic Chains, introduced by \citet{tsao77} and demonstrates that people omit pronouns when a certain kind of discourse salience is high. We show that this notion of salience is robust across various choices of language models, however, locality (\textit{i.e.} clause recency) seems to be a key requirement.

The proposed formalization leverages the idea that verbs predicated on the same story-character exhibit discourse coherence \citep{huang1984distribution, huang1994syntax, li1979third}. Figure \ref{Fig:example} shows a literary example where the same character, the narrator, is explicitly referred to once using an explicit pronoun ``wo''. After that, the pronoun is dropped. The list of predicates (shown in red) applying to the narrator in examples (1) - (3) is [``draw'', ``lose'', ``draw'']. When faced with another omitted pronoun in example (4), the fact that the predicate is also ``draw'' supports the interpretation that the omitted element refers to the narrator. This is because ``draw'' is similar to the history verbs ``draw'' and ``lose'' which were predicated of this same character earlier in the discourse. In this short example, there are other entities such as ``grownups'' and the ``boa constrictor'', but their verb histories make them less plausible as candidate referents of the omitted pronoun.

\begin{figure*}
    \centering
    \includegraphics[scale = 0.7]{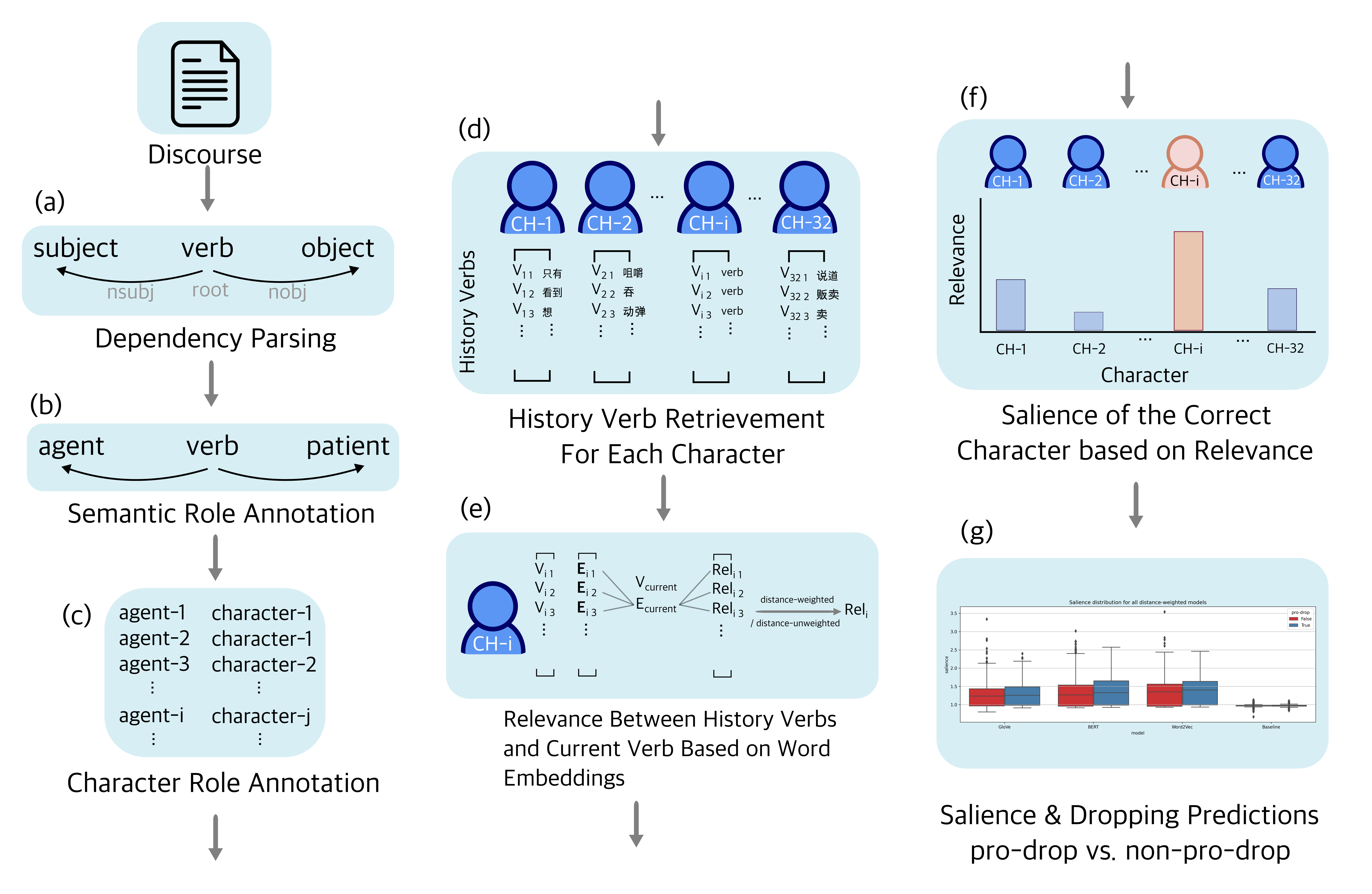}
    \caption{Analysis steps adopted in this study: (a) Grammatical subjects and objects of each main verb are identified via dependency parsing on the whole story discourse of The Little Prince (See a sentence example from Table \ref{Tab:annotation_table}, columns ``S'', ``V'', ``O''); (b) Semantic role annotation: for all the subjects and objects, annotate their semantic roles as \textsc{Agent} or \textsc{Patient} (See Table \ref{Tab:annotation_table} column ``V-agent'' and ``V-patient''); (c) Character role annotation: assign story character roles to the entities, see character occurrences in Table \ref{Tab:character_distribution}, and Table \ref{Tab:annotation_table} column ``character''; (d) History verb retrieval for each story character: for each story character, tabulate the verbs that are its main verbs being used in the discourse (See example Table \ref{Tab:character_verb_case}); (e) Relevance between history verbs and a current verb: for each current verb, calculate its relevance to the history verbs, and sum with or without their distance weight (See Table \ref{Tab:relevance_regressor} and \ref{Tab:relevance_result_case}); (f) Salience of the correct character: for each verb, calculate how ``salient'' the correct character is compared to all other characters (See example Table \ref{Tab:salience-case}); (g) Group test between \textit{pro}-drop verbs vs. non-\textit{pro}-drop verbs, and apply logistic regression to test predictability of character salience on dropping behavior (See group results in Table \ref{Tab:salience-result} and Figure \ref{Fig:salience}).}
    \label{Fig:intro}
\end{figure*}

In this paper, we use representations from three neural language models to quantify character-verb usage continuity in a literary discourse, and calculate salience values for each of 32 possible characters at the site of each omitted pronoun. Figure \ref{Fig:intro} summarizes the analytical steps of this process. Our contributions are as follows: (1) We provide a numerical description of the topic chain continuity. (2) We elaborate on the role of verbs in resolving omitted pronouns. (3)  We show that verb similarity and clause range offer reliable clues about the referent of the omitted pronoun.

\section{Related Work}

Various linguistic theories point to discourse coherence as a factor that enables or encourages \textit{pro}-drop. One of these is Tsao's (\citeyear{tsao77}) notion of Topic Chain. As reviewed in \citet{pu2019routledge}, a topic~chain is a sequence of clauses sharing an identical topic that occurs overtly in one of the clauses. Topic Chains may cross several sentences and even paragraphs \citep{li2004topic}. The multiclausal aspect of Topic Chains supports long-distance coreference \citep{sun2019integration}. Taking a dynamic perspective, \citet{pu2019zero} suggests that a topic chain ``encodes a referent that is cognitively most accessible at the moment of discourse production, as enhanced by maximum discourse coherence of topic continuity and thematic coherence''.


We conceptualize accessibility in Pu's sense as the relative salience of a story character that participates in a chain of predications. Instead of focusing on named entities, we form the chain based on the verbs in the preceding discourse.

\begin{table}
\begin{CJK}{UTF8}{gbsn} 
\centering
\resizebox{\columnwidth}{!}{%
\begin{tabular}{|c|c|c|c|c|c|c|c|}
\hline
\textbf{ID} & \textbf{word}                                        & \textbf{S} & \textbf{V} & \textbf{O} & \textbf{V-agent}                                  & \textbf{V-patient}                                 & \textbf{character}  \\ \hline
56          & \begin{tabular}[c]{@{}c@{}}这些 (these)\end{tabular}    &            &            &            &                                                   &                                                    &                                             \\ \hline
57          & \begin{tabular}[c]{@{}c@{}}蟒蛇 (boa)\end{tabular}     & True         &            &            &                                                   &                                                    & ch2\_boa                                    \\ \hline
58          & \begin{tabular}[c]{@{}c@{}}把 (BA)\end{tabular}       &            &            &            &                                                   &                                                    &                                             \\ \hline
59          & \begin{tabular}[c]{@{}c@{}}它们 (them)\end{tabular}    &            &            &          &                                                   &                                                    &                                             \\ \hline
60          & \begin{tabular}[c]{@{}c@{}}的 (DE)\end{tabular}       &            &            &            &                                                   &                                                    &                                             \\ \hline
61          & \begin{tabular}[c]{@{}c@{}}猎获物 (prey)\end{tabular}   &            &            & True        &                                                   &                                                    &                                    \\ \hline
62          & \begin{tabular}[c]{@{}c@{}}不 (not)\end{tabular}      &            &            &            &                                                   &                                                    &                                             \\ \hline
63          & \begin{tabular}[c]{@{}c@{}}加 (with)\end{tabular}      &            &            &            &                                                   &                                                    &                                             \\ \hline
64          & \begin{tabular}[c]{@{}c@{}}咀嚼 (chew)\end{tabular}    &            & True         &            & \begin{tabular}[c]{@{}c@{}}57 (boa)\end{tabular} & \begin{tabular}[c]{@{}c@{}}61 (prey)\end{tabular} &                                             \\ \hline
65          & \begin{tabular}[c]{@{}c@{}}地 (DI)\end{tabular}       &            &            &            &                                                   &                                                    &                                             \\ \hline
66          & \begin{tabular}[c]{@{}c@{}}囫囵 (roughly)\end{tabular} &            &            &            &                                                   &                                                    &                                             \\ \hline
67          & \begin{tabular}[c]{@{}c@{}}吞 (swallow)\end{tabular} &            & True          &            & \begin{tabular}[c]{@{}c@{}}57 (boa)\end{tabular} & \begin{tabular}[c]{@{}c@{}}61 (prey)\end{tabular} &                                             \\ \hline
68          & \begin{tabular}[c]{@{}c@{}}下 (down)\end{tabular}     &            &            &            &                                                   &                                                    &                                             \\ \hline

\end{tabular}}
\caption{Dependency structure and semantic role annotation table. An annotation example for the sentence ``These boas swallow their prey without chewing.'' The verbs ``chew'' and ``swallow'' are located as verbs in the column \textit{V}. Token indices for each verb's Agent and/or Patient are annotated in the columns \textit{V-agent} and \textit{V-patient} respectively, and the character roles they are referring to are annotated in the column \textit{character}. }

\label{Tab:annotation_table}
\end{CJK} 
\end{table}

\section{Method}

\subsection{Discourse Material}
The discourse material used in this study is a Chinese translation \citep{xiaowangzi} of
Saint-Exup\'ery's \textit{The Little Prince}. It contains 2802 clauses and 16010 words, and the word tokenization was manually checked by native Chinese speakers. 

\subsection{Dependency Structure Retrieval and Semantic Role Annotation}
\label{Sec:annotation_dependency}
We manually annotate the semantic roles Agent and Patient for each verb using dependency analyses provided by Stanza \citep{qi2020stanza} and part of speech tags provided by spaCy. For most cases in the discourse, subjects are acting as agents whereas objects are acting as patients, but there are 494 exceptions (\textit{i.e.} 218 Agents are acting as Objects, and 276 Patients are acting as Subjects) such as passives, the -BA(`把') construction, the relative clause -DE(`的') construction \textit{etc.} that call for our manual annotation (See Chapter 28 and 32 in The Oxford Handbook of Chinese Linguistics \citep{wang2015oxford} regarding these constructions).

The textual antecedents of each agent and patient are separately annotated manually. As shown in Table \ref{Tab:annotation_table}, the sentence meaning ``These boas swallow their prey without chewing'' has the following annotations: verbs annotated in column \textit{V}; verbs' agents and patients annotated correspondingly in column \textit{V-agent} and \textit{V-patient}; pronouns or named entities' character roles are annotated in column \textit{character}. As described below in Section \ref{Sec:character-verb table}, information about characters in particular semantic roles can be used to form a dynamic usage table, reifying Pu's view of Topic Chains.

\subsection{\textit{Pro}-drop Annotation}
\label{Sec:pro-drop annotation}
Omitted subjects and objects are manually resolved using numerical indices from 1 to 32. As shown in Appendix Table \ref{Tab:annotation_result}, 422 Agents and 16 Patients are found omitted in the discourse, and in the following analyses, we focus on just story characters in the Agent semantic role.

\subsection{Dynamic Character-Verb Usage Table}
\label{Sec:character-verb table}


Based on the dependency annotation table, the verbs used for each character are extracted and entered in a second table, the Character-Verb Usage Table (See example in Appendix Table \ref{Tab:character_verb_case}). This table includes the following features: (1) \textit{verb}, the original text of the verb; (2) \textit{verb\_id}, the index of the verb in the whole discourse; (3) \textit{agent/patient\_character}, the verb's agent or patient story character; (4) \textit{pro\_drop}, whether the verb has \textit{pro}-drop; (5) \textit{ch[1-32]\_prev\_verbs}, for characters 1 through 32, their corresponding previous verbs and indexes are stored as lists.

The dynamic character-verb usage table includes the previous verbs for each story character until a ``current verb'', and this indicates the verb usage history of each character. By transforming these verb usage histories into numerical vectors, it is possible to use a simple notion of similarity to formalize discourse coherence.

\subsection{History-verb and Current-verb Relevance}
\label{Sec:verb_relevance}
The idea behind comparing the history verbs and the current verb for each story character is to calculate a numerical similarity level between the current verb and preceding verbs that are part of one or another Topic Chain. Inspired by Sperber and Wilson (\citeyear{sperber1986relevance}), we define a quantity called Relevance, a time-weighted function of vector similarity with preceding predicates. The Relevance evaluation process adopt three types of word embeddings (See Section \ref{Sec:word_embedding} for details), and steps for the evaluation are introduced in Section \ref{Sec:relevance_ev}.   


\subsubsection{Word Embeddings Methods}
\label{Sec:word_embedding}

Word embeddings allow each word to be mapped to a single point in a vector space. Under the Distributional Hypothesis (see \textit{e.g.} \citealp{lenci2018distributional}), words with similar meanings should be closer in vector space (for a textbook introduction, see \citealp{pilehvar2020embeddings}). We use this idea to calculate the similarity between the main verb of an omitted pronoun and the verb chains of story characters that might serve as that omitted pronoun's referent. 

We use three types of word embeddings: GloVe, BERT, and Word2Vec. The GloVe model \citep{pennington2014glove} learns word embedding from the term co-occurrence matrix by minimizing the reconstruction error. GloVe has a large context window, which allows it to capture longer-term dependency features. The BERT model \citep{devlin2018bert} consists of multi-layer bidirectional transformer encoders. BERT is trained on two unsupervised tasks: predict masked tokens, and predict the next sentence, and the BERT embeddings reflect contextual corpus features. Word2Vec is a prediction-based model \citep{mikolov2013efficient, mikolov2013distributed}, and the word embeddings used in this study \citep{li-etal-2018-analogical} were trained on a Skip-Gram with Negative Sampling (SGNS) model. All word embeddings we used were trained on large Chinese corpora, and contain contextual word knowledge that carries semantic, syntactic, and pragmatic features. Among these three word embedding models, BERT can provide contextualized features of the language compared to the others due to the tasks and processes it has been trained on.

In this study, BERT and GloVe models are applied with spaCy\footnote{\href{https://spacy.io/models/zh}{https://spacy.io/models/zh}}, and Word2Vec model is applied with pretrained Chinese Word Vectors\footnote{\href{https://github.com/Embedding/Chinese-Word-Vectors}{https://github.com/Embedding/Chinese-Word-Vectors}} \citep{li-etal-2018-analogical}. A baseline model with 300-dimension random value vectors is adopted to calculate the baseline relevance as compared to the other word embedding models.

\textbf{The GloVe word embeddings} are obtained from the \textit{zh\_core\_web\_lg} model in spaCy. The GloVe model \citep{pennington2014glove} relies on word co-occurrence in the training corpus, and considers the ratios of word-word co-occurrence probabilities to encode semantic information. The model in spaCy was trained on OntoNotes 5, CoreNLP Universal Dependencies Converter, and Explosion fastText Vectors. It has 500,000 unique vectors with a dimension size of 300. We obtained the word vectors by searching up the Chinese word in the word dictionary.

\textbf{The BERT word embeddings} are retrieved from the \textit{zh\_core\_web\_trf} model in spaCy. This transformer model was trained on OntoNotes 5, CoreNLP Universal Dependencies Converter, and bert-base-chinese. The word embedding vectors were obtained by grouping every 50 words in the discourse, and the model inputs were the 50 words combined as a string (with space between the words). The dimension of the BERT word embedding is 768. If there were more than 1 character in a word, their vectors' mean value was used as the word embedding for the whole word. For example, the word ``只有'' 's embedding was calculated by averaging its subwords' embedding vectors of ``只'' and ``有''.

\textbf{The Word2Vec word embeddings} were pretrained on Word2Vec model with a large Chinese corpus containing data from Baidu Netdisk (22.6G), and the vector dimension is 300 \citep{li-etal-2018-analogical}.

\textbf{Baseline Word Vectors} were 300-dimension vectors generated randomly in the range -1 to 1. The same analysis steps are applied to this model as a baseline.

\subsubsection{Relevance evaluation}
\label{Sec:relevance_ev}
The relevance between history verbs and current verbs is calculated based on their word embedding similarities (see Section \ref{Sec:word_embedding} for details). At the same time, a weight decay function is applied to the influence of each history verb based on its distance to the current verb, and the function used here is a vanilla value decreasing function (see Equation \ref{eq1}), in which $\omega$ refers to the weight applying on the similarity, \textit{d} refers to the clause distance between the verbs being compared, and\textit{ j}, \textit{k} are the clause numbers the verbs are in:

\begin{equation} \label{eq1}
\begin{split}
\omega(j, k) = 1/(d + 1) \\
d = \left | j - k \right |
\end{split}
\end{equation}

In this study, the ``word embedding similarity'' method is realized by calculating the Cosine Similarity between two word embedding vectors. As shown in Equation \ref{eq2}, \textit{$v_{prev}$} refers to a word embedding vector of a previous verb, and \textit{$v_{curr}$} refers to the one for the current verb:


\begin{equation} \label{eq2}
R(v_{prev}, v_{curr}) = \frac{v_{prev} \cdot v_{curr}}{||v_{prev} || ||v_{curr} ||}
\end{equation}

Therefore, the clause-distance-weighted similarity between history verbs and the current verb is shown as Equation \ref{eq3}, in which \textit{n} refers to the number of verbs in the history verb list for a character, and \textit{cl\_prev\_i} and \textit{cl\_curr} refer to the clause numbers that the previous verb and the current verb are in correspondingly.

\begin{equation} \label{eq3}
\begin{split}
R_{weighted}([v_{prev\_1}, ..., v_{prev\_n}], v_{curr}) = \\ \sum_{i = 1}^{n}\omega(cl\_prev\_i, cl\_curr) * R(v_{prev\_i}, v_{curr})
\end{split}
\end{equation}

Via Equation \ref{eq3}, for a current verb, each story character has a corresponding relevance value: if the value is higher, the distance-weighted word embedding similarity between history verbs and current verb is higher; and vice versa. 

Appendix Table \ref{Tab:character_verb_case} shows an example of a verb and the history verbs for characters 1 through 32. The GloVe, BERT, Word2Vec, and Baseline embeddings are used to calculate the average relevance of the history verbs to each current verb for each story character.

Regressors obtained from relevance evaluation introduced in this section are shown in Table \ref{Tab:relevance_regressor}. The average similarity is calculated following Equation \ref{eq2} and \ref{eq3}. Both distance-weighted and distance-unweighted relevance are explored to see whether clause distance would play a role.

\begin{table}[!ht]
\begin{CJK}{UTF8}{gbsn} 
\centering
\resizebox{\columnwidth}{!}{%
\begin{tabular}{|c|c|l|}
\hline
\textbf{Regressor Number} & \textbf{Regressor Name} & \multicolumn{1}{c|}{\textbf{Regressor Meaning}}                                                                     \\ \hline
1                         & verb                    & \begin{tabular}[c]{@{}l@{}}the verb in the discourse acting \\ as a main verb of a clause\end{tabular}              \\ \hline
2                         & verb-id                 & \begin{tabular}[c]{@{}l@{}}the word order id of this verb \\ in the original discourse\end{tabular}                 \\ \hline
3                         & agent-character         & \begin{tabular}[c]{@{}l@{}}the story character referred \\ by the agent of the verb\end{tabular}                    \\ \hline
4                         & pro-drop                & \begin{tabular}[c]{@{}l@{}}whether this agent is dropped \\ in the discourse\end{tabular}                           \\ \hline
5 - 36                    & ch\{1-32\}-prev-verbs    & \begin{tabular}[c]{@{}l@{}}the previous verbs used by \\ each story character till the \\ current verb\end{tabular} \\ \hline
37 - 68                   & rel-glove-ch\{1-32\}     & \begin{tabular}[c]{@{}l@{}}relevance obtained by \\ GloVe word embeddings\end{tabular}                              \\ \hline
69 - 100                   & rel-bert-ch\{1-32\}      & \begin{tabular}[c]{@{}l@{}}relevance obtained by \\ BERT word embeddings\end{tabular}                               \\ \hline
101 - 132                   & rel-word2vec-ch\{1-32\}     & \begin{tabular}[c]{@{}l@{}}relevance obtained by \\ Word2Vec word embeddings\end{tabular}                       \\ \hline
133 - 164                   & rel-baseline-ch\{1-32\}     & \begin{tabular}[c]{@{}l@{}}relevance obtained by \\ Baseline word vectors\end{tabular}                       \\ \hline
\end{tabular}}
\caption{Regressors obtained after the relevance calculation}
\label{Tab:relevance_regressor}
\end{CJK} 
\end{table}

As shown in Appendix Table \ref{Tab:relevance_result_case}, the relevance calculation results of the last verb are presented as an example.

\subsection{Character Salience}
\label{character_salience}

With the relevance between history-current verbs computed as described in the previous section, we have a similarity value for each story character to the current verb. This character salience value refers to whether a story character stands out compared to other candidate characters. The salience level function is described in Equation \ref{eq4}. In Equation \ref{eq4}, \textit{k} refers to character\_k, and the relevance values were calculated based on its history-current verbs by Equation \ref{eq3}. 


\begin{equation} 
\label{eq4}
S(k) =\frac{\sum_{i=1}^{n}\left ( \frac{R_{weighted}(k) + 1}{R_{weighted}(i) + 1 } \right )}{n + 1}
\end{equation}

\begin{table*}[t]
\begin{adjustbox}{width=2\columnwidth,center}
\centering
\begin{tabular}{cccccccccc}
\hline
\multicolumn{2}{c}{}                                                                                         & \multicolumn{8}{c}{\textbf{\begin{tabular}[c]{@{}c@{}}Correct character salience\\ pro-drop \textgreater non-pro-drop\\ (n = 422)\end{tabular}}}                                                                    \\ \hline
\multicolumn{2}{c}{\textbf{Candidates' Range}}                                                               & \multicolumn{2}{c}{\textbf{range = all}} & \multicolumn{2}{c}{\textbf{range \textless 10 clause}} & \multicolumn{2}{c}{\textbf{range \textless 20 clause}} & \multicolumn{2}{c}{\textbf{range \textless 30 clause}} \\ 
\multicolumn{2}{c}{}                                                                                         & \textit{t-value}    & \textit{p-value}   & \textit{t-value}           & \textit{p-value}          & \textit{t-value}           & \textit{p-value}          & \textit{t-value}           & \textit{p-value}          \\ \hline
\multirow{4}{*}{\textbf{\begin{tabular}[c]{@{}c@{}}Distance-\\ Weighted\end{tabular}}}   & \textbf{GloVe}    & 49090.319           & 0.063              & 51137.593                  & 0.012*                    & 52598.233                  & 0.003**                   & 52121.241                  & 0.004**                   \\ \cline{2-10} 
                                                                                         & \textbf{BERT}     & 50555.45            & 0.023*             & 45310.076                  & 0.029*                    & 52105.854                  & 0.005**                   & 51582.819                  & 0.008**                   \\ \cline{2-10} 
                                                                                         & \textbf{Word2Vec} & 50358.954           & 0.025*             & 51268.800                  & 0.011*                    & 52747.81                   & 0.002**                   & 52246.569                  & 0.004**                   \\ \cline{2-10} 
                                                                                         & \textbf{Baseline} & 44656.318           & 0.496              & 44737.336                  & 0.483                     & 49199.853                  & 0.060                     & 47875.291                  & 0.134                     \\ \hline
\multirow{4}{*}{\textbf{\begin{tabular}[c]{@{}c@{}}Distance-\\ Unweighted\end{tabular}}} & \textbf{GloVe}    & 39345.494           & 0.959              & 44384.169                  & 0.531                     & 43818.383                  & 0.606                     & 43837.85                   & 0.604                     \\ \cline{2-10} 
                                                                                         & \textbf{BERT}     & 42867.41            & 0.724              & 45310.076                  & 0.411                     & 45187.343                  & 0.425                     & 45220.75                   & 0.421                     \\ \cline{2-10} 
                                                                                         & \textbf{Word2Vec} & 40865.782           & 0.898              & 45236.126                  & 0.420                     & 44672.755                  & 0.494                     & 44630.117                  & 0.498                     \\ \cline{2-10} 
                                                                                         & \textbf{Baseline} & 43149.674           & 0.690              & 45940.625                  & 0.330                     & 46398.831                  & 0.275                     & 45552.563                  & 0.377                     \\ \hline
\end{tabular}
\end{adjustbox}
\caption{Single-sided nonparametric two-sample Wilcoxon test between \textit{pro}-drop and non-\textit{pro}-drop salience values among three word embedding models and the baseline model: With candidates included as all candidates, candidates within 10 clauses, 20 clauses, and 30 clauses.}
\label{Tab:salience-result}
\end{table*}

\begin{table*}[t]
\begin{adjustbox}{width=2\columnwidth,center}
\centering
\begin{tabular}{cccccc}
\hline
\multicolumn{2}{c}{}                                                                                         & \multicolumn{4}{c}{\textbf{\begin{tabular}[c]{@{}c@{}}Logistic Regression Model \\ Pro-drop Prediction Accuracy\end{tabular}}}      \\ \hline
\multicolumn{2}{c}{\textbf{Candidates' Range}}                                                               & \textbf{range = all} & \textbf{range \textless 10 clause} & \textbf{range \textless 20 clause} & \textbf{range \textless 30 clause} \\ \hline
\multirow{4}{*}{\textbf{\begin{tabular}[c]{@{}c@{}}Distance-\\ Weighted\end{tabular}}}   & \textbf{GloVe}    & \textbf{0.518}       & \textbf{0.535}                     & \textbf{0.527}                     & \textbf{0.539}                     \\ \cline{2-6} 
                                                                                         & \textbf{BERT}     & \textbf{0.538}       & \textbf{0.532}                     & \textbf{0.536}                     & \textbf{0.546}                     \\ \cline{2-6} 
                                                                                         & \textbf{Word2Vec} & \textbf{0.534}       & \textbf{0.535}                     & \textbf{0.537}                     & \textbf{0.552}                     \\ \cline{2-6} 
                                                                                         & \textbf{Baseline} & 0.497                & 0.489                              & 0.495                              & 0.498                              \\ \hline
\multirow{4}{*}{\textbf{\begin{tabular}[c]{@{}c@{}}Distance-\\ Unweighted\end{tabular}}} & \textbf{GloVe}    & \textbf{0.524}       & 0.487                              & 0.490                              & 0.485                              \\ \cline{2-6} 
                                                                                         & \textbf{BERT}     & 0.493                & 0.488                              & 0.492                              & 0.482                              \\ \cline{2-6} 
                                                                                         & \textbf{Word2Vec} & \textbf{0.514}       & 0.485                              & 0.482                              & 0.473                              \\ \cline{2-6} 
                                                                                         & \textbf{Baseline} & 0.485                & 0.485                              & 0.485                              & 0.485                              \\ \hline
\end{tabular}
\end{adjustbox}
\caption{\textit{Pro}-drop prediction accuracy results of the Logistic Regression model from three word embedding models and one baseline model: salience value calculated based on all previous clauses and ranged clauses.}
\label{Tab:accuracy-result}
\end{table*}

\subsubsection{Ranged Character Salience}
Instead of taking all 32 story characters as candidates for the salience value calculation, the ranged candidates' salience compares the correct character's accumulated relevance value to the ones within a certain number of clauses. We consider candidates within 10, 20, and 30 clauses for this ranged salience.

\subsection{\textit{Pro}-drop Prediction}
\label{prediction}
With the correct story character's salience level for each verb in the annotated discourse, we apply a logistic regression model to predict \textit{pro}-drop based on the salience level. The sample sizes are chosen by the size of the smaller group (\textit{i.e.} \textit{pro}-drop), and the chosen processes are repeated 100 times to obtain the average accuracy level.

\section{Results}
In this study, the following analyses are applied to The Little Prince discourse to explore the effect of verb continuity on the \textit{pro}-drop phenomenon: (1)  relevance between history and current verbs for all story characters (with three types of word embeddings applied); (2) character salience of the correct character, and (3) correct character salience group comparison, and its predictability on \textit{pro}-drop in the discourse. The following sections describe the results of (2) and (3), and (1) is an intermediate step introduced in Section \ref{Sec:verb_relevance}.




\begin{figure*}[ht]
    \centering
    \scalebox{0.75}{\includegraphics{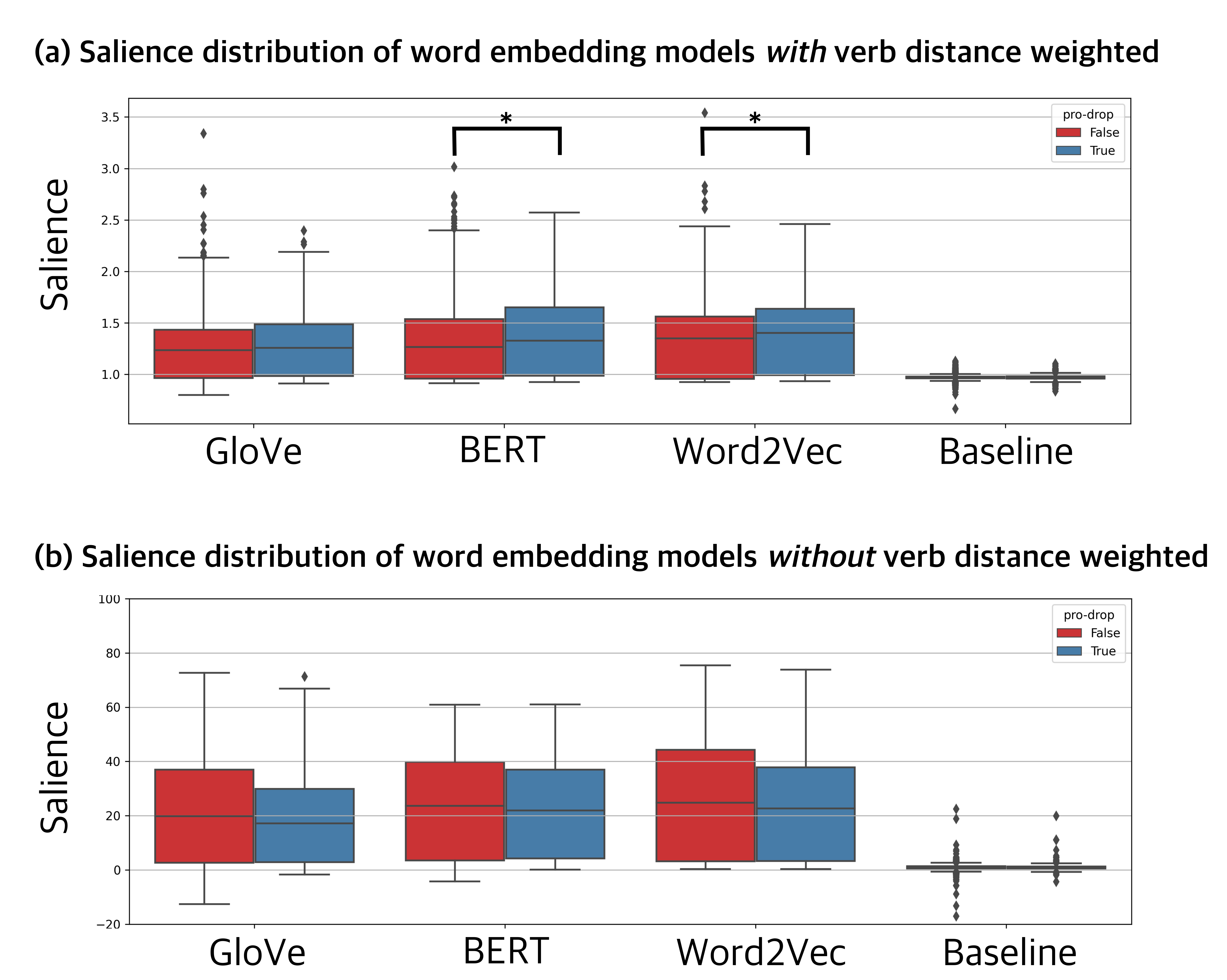}}
    \caption{Salience distributions from the word embedding models: GloVe, BERT, Word2Vec, and Baseline. (a) Salience distribution based on distance-weighted models; (b) Salience distribution based on distance-unweighted models. The blue boxplots are \textit{pro}-drop salience cases, and the red ones are non-\textit{pro}-drop. The BERT and Word2Vec models show significant \textit{pro}-drop > non-\textit{pro}-drop effect, and GloVe model shows marginally significant result (See detailed Wilcoxon tests results in Table \ref{Tab:salience-result}).}
    \label{Fig:salience}
\end{figure*}

\subsection{Character Salience: \textit{Pro}-drop vs. Non-\textit{pro}-drop}

The correct story character's salience compared to all other characters was calculated following Equation \ref{eq4}. For each verb, we calculated a salience value for the correct story character. See Appendix Table \ref{Tab:salience-case} for an example of the salience values of the last verb.

The distributions for the salience value obtained from three word embedding models and one baseline model are shown in Figure \ref{Fig:salience}.

Single-sided nonparametric two-sample Wilcoxon Tests are carried out between \textit{pro}-drop and non-\textit{pro}-drop character salience of all three word embedding models and the baseline model. 422 cases are randomly selected from non-\textit{pro}-drop salience values to match the size of the \textit{pro}-drop ones, and this process is repeated 1000 times to gain the average statistic values. The test results are shown in Table \ref{Tab:salience-result}. For distance-weighted models, BERT and Word2Vec show significant results (\textit{p} $<$ 0.05), and GloVe show marginally significant result (\textit{p} = 0.063). For distance-unweighted models, none of them show significant results. The Wilcoxon test results based on ranged salience are shown in Table \ref{Tab:salience-result} in columns ``range $<$ 10/20/30 clauses'' as compared to the non-ranged results in ``range = all''. As shown in the table, the effects of ``\textit{pro}-drop $>$ non-\textit{pro}-drop'' on correct character salience tend to be larger when the salience is calculated based on ranged clauses. The Baseline model shows null effects on both distance-weighted and distance-unweighted models for all the ranged cases. As shown in Figure \ref{Fig:salience}, the boxplots are consistent with the Wilcoxon tests.

\subsection{Logistic Regression: Predict \textit{Pro}-drop with Character Salience}

With the salience values described in the previous section, the logistic regression model is applied to examine the effect of salience on \textit{pro}-drop. 75\% of the data are used as the training set, and 25\% of the data are used as the testing set. See the prediction accuracy results in Table \ref{Tab:accuracy-result} based on salience values obtained from all-ranged and clause-ranged clauses. As shown in Table \ref{Tab:accuracy-result}, except for the baseline model, all the distance-weighted language models' results show above chance ( $>$ 50\%) accuracy. As for distance-unweighted language models, only GloVe and Word2Vec show above chance results on all-ranged predictions. Similar to the ``range-effect'' shown in the previous section, it can be seen from the prediction results as well that ranged clauses' prediction accuracies tend to be slightly higher than non-ranged results.

\section{Discussion}

The main findings of this study are: (1) Compared with overtly expressed subjects, omitted subjects have higher verb-usage continuity. In this respect, they stand out among other story characters; (2) Clause distance plays a role in contextual information strength: With clause distance weighted, the \textit{pro}-drop $>$ non-\textit{pro}-drop salience effects are statistically significant; (3) Constraining the range of candidates by clause recency appears to strengthen these effects.

These results validate Topic Chain theory \citep{tsao77} by showing how verbs contribute to the discourse coherence that omitted pronouns depend on. The ``ranged'' recency effect indicates that local contextual coherence might play a more important role than whole-discourse-level coherence. This recency effect may also explain the better performance obtained by BERT and Word2Vec compared to GloVe, since GloVe word embeddings are obtained from discourse-level word co-occurrence statistical features, and BERT and Word2Vec are trained on comparably smaller scale contextual information.

It shall be noted that verb-usage continuity is not the only factor that conditions \textit{pro}-drop. Other factors, including nonverbal lexical information and syntactic patterns \textit{e.g.} with conjunctions, also support discourse coherence \citep{hasan1976cohesion}. In this light, it is remarkable that one factor on its own, verb-usage continuity, yields above-chance accuracy in predicting \textit{pro}-drop.


\section{Conclusion}
This study quantifies character-verb usage continuity as an aspect of discourse that helps comprehenders resolve omitted pronouns. Omitted pronouns tend to show higher verb usage consistency compared to pronounced entities, and this effect is strengthened by clause recency. 

\section*{Acknowledgements}
This material is based upon work supported by the National Science Foundation under Grant No. 1903783.

\bibliography{qp2}
\bibliographystyle{acl_natbib}

\clearpage
\setcounter{table}{0}
\renewcommand{\thetable}{A\arabic{table}}

\appendix
\section{Appendix}
\label{sec:appendix}

\begin{table}[!h]
\centering
\resizebox{0.95\textwidth}{!}{
\fontsize{8}{11}\selectfont
\begin{tabular}{|c|c|c|}
\hline
\textbf{Story character} & \textbf{Annotation Label} & \textbf{Number of Occurrence} \\ \hline

the little prince        & ch4\_prince             & 676                           \\ \hline
the story teller         & ch1\_storyteller        & 356                           \\ \hline
the rose                 & ch12\_rose              & 166                           \\ \hline
the king                 & ch18\_king              &  71                           \\ \hline
the fox                  & ch28\_fox               &  67                           \\ \hline
the planet               & ch11\_planet            &  62                           \\ \hline
the lamplighter          & ch23\_lighter           &  54                           \\ \hline
the sheep                & ch5\_sheep              &  48                           \\ \hline
the geologist            & ch24\_geologist         &  41                           \\ \hline
the grownups             & ch3\_grownups           &  39                           \\ \hline
the snake                & ch26\_snake             &  39                           \\ \hline
the businessman          & ch22\_shiyejia          &  37                           \\ \hline
readers                  & ch8\_audience           &  30                           \\ \hline
the volcano              & ch17\_volcano           &  22                           \\ \hline
the baobab               & ch9\_tree               &  20                           \\ \hline
the drunk man            & ch21\_drunk             &  18                           \\ \hline
the conceited man        & ch20\_xurong            &  16                           \\ \hline
the travelers            & ch31\_traveler          &  15                           \\ \hline
the seed                 & ch10\_grass             &  13                           \\ \hline
the explorer             & ch25\_explorer          &  13                           \\ \hline
the red-faced man        & ch14\_redface           &  11                           \\ \hline
the boa                  & ch2\_boa                &  10                           \\ \hline
the switch man           & ch29\_switcher          &  10                           \\ \hline
the astronomer           & ch6\_universescholar    &   7                           \\ \hline
the echo                 & ch27\_echo              &   5                           \\ \hline
the tiger                & ch15\_tiger             &   5                           \\ \hline
the drafts               & ch16\_wind              &   4                           \\ \hline
the train                & ch30\_train             &   4                           \\ \hline
the merchant             & ch32\_merchant          &   4                           \\ \hline
the children             & ch13\_kids              &   3                           \\ \hline
the general              & ch19\_general           &   3                           \\ \hline
the ruler                & ch7\_ruler              &   1                           \\ \hline

\end{tabular}
}
\caption{The number of occurrence of each \\ character in the annotated discourse}

\label{Tab:character_distribution}
\end{table}

\begin{table*}
\centering
\resizebox{1\columnwidth}{!}{%
\begin{tabular}{|c|c|c|}
\hline
                                                                              & \textbf{Agent} & \textbf{Patient} \\ \hline
\textbf{\textit{pro}-drop}                                                             & 422            & 16               \\ \hline
\textbf{non-\textit{pro}-drop}                                                         & 2032           & 1329             \\ \hline
\textbf{total number} & 2454           & 1345             \\ \hline
\end{tabular}
}
\caption{Distribution of annotated Agents and Patients in the whole discourse.}
\label{Tab:annotation_result}
\end{table*}

\begin{table*}
\begin{CJK}{UTF8}{gbsn} 
\centering
\resizebox{1.8\columnwidth}{!}{%
\begin{tabular}{|c|l|}
\hline
\textbf{verb}             & 回来                                                      \\ \hline
\textbf{verb\_id}         & 16008                                                    \\ \hline
\textbf{agent\_character} & ch4                                                     \\ \hline
\textbf{pro\_drop}        & False                                                   \\ \hline
\textbf{ch1\_prev\_verbs} &  [只有, 看到, 想, 用, 画, 画, 让, 画,...] \\ \hline
\textbf{ch2\_prev\_verbs} &   [咀嚼, 吞, 动弹, 消化, 消化, 开, 闭, 闭,...]\\ \hline
\textbf{ch3\_prev\_verbs} &   [理解, 看, 懂, 需要, 解释, 劝, 靠, 弄,...]\\ \hline
\textbf{ch4\_prev\_verbs} &   [朝, 望, 出现, 给, 像, 没有, 像, 干,...]\\ \hline
\textbf{ch5\_prev\_verbs} &   [病, 需要, 像, 睡, 去, 用, 跑, 跑,...]\\ \hline
\textbf{...} &   ...            \\ \hline
\textbf{ch30\_prev\_verbs} &   [运载, 发, 往, 朝着, 开, 过] \\ \hline
\textbf{ch31\_prev\_verbs} &   [寻找, 回来, 满意, 住, 追随, 追随, 睡觉, 打哈欠,...]\\ \hline
\textbf{ch32\_prev\_verbs} &   [说道, 贩卖, 卖, 说]\\ \hline
\end{tabular}
}
\caption{Example of Verb-Character table. (See a translation of this table in Table \ref{Tab:trans})}
\label{Tab:character_verb_case}
\end{CJK} 
\end{table*}

\begin{table*}
\begin{CJK}{UTF8}{gbsn} 
\centering
\resizebox{2\columnwidth}{!}{%
\begin{tabular}{|c|l|}
\hline
\textbf{verb}             & come back                                                      \\ \hline
\textbf{verb\_id}         & 16008                                                    \\ \hline
\textbf{agent\_character} & ch4                                                     \\ \hline
\textbf{pro\_drop}        & False                                                   \\ \hline
\textbf{ch1\_prev\_verbs} &  [have, see, want, use, draw, draw, let, draw,...] \\ \hline
\textbf{ch2\_prev\_verbs} &   [chew, swallow, move, digest, digest, open, close, close,...]\\ \hline
\textbf{ch3\_prev\_verbs} &   [understand, see, understand, need, explain, advise, lean, play,...]\\ \hline
\textbf{ch4\_prev\_verbs} &   [turn, watch, show up, give, alike, (not) have, alike, do,...]\\ \hline
\textbf{ch5\_prev\_verbs} &   [sick, need, alike, sleep, go, use, run, run,...]\\ \hline
\textbf{...} &   ...            \\ \hline
\textbf{ch30\_prev\_verbs} &   [carry, send, go, turn, drive, pass] \\ \hline
\textbf{ch31\_prev\_verbs} &   [look up, come back, satisfy, live, follow, follow, sleep, yawn,...]\\ \hline
\textbf{ch32\_prev\_verbs} &   [speak, sell, sell, say]\\ \hline
\end{tabular}
}
\caption{Translation of Table \ref{Tab:character_verb_case}: Example of Verb-Character table.}
\label{Tab:trans}
\end{CJK} 
\end{table*}

\begin{table*}
\begin{CJK}{UTF8}{gbsn} 
\centering
\resizebox{1.8\columnwidth}{!}{%
\begin{tabular}{|l|l|}
\hline
\textbf{Relevance Regressor} & \textbf{(Non-weighted relevance, Weighted relevance)} \\ \hline
rel\_glove\_ch1              & (81.89066125531684, 0.32419914580071807)                \\ \hline
rel\_glove\_ch2              &  (1.8756812506219913, 0.001503683756709864)          \\ \hline
...           & ...            \\ \hline
rel\_glove\_ch32              &   (0.8230171383397842, 0.001262691669193839)       \\ \hline
rel\_bert\_ch1               &    (176.59183087820725, 0.6119750732174682)  \\ \hline
rel\_bert\_ch2               &    (4.919826668243348, 0.0027848581443943223)        \\ \hline
...           & ...            \\ \hline
rel\_bert\_ch32               &   (0.867459723760406, 0.001329274033713714) \\ \hline
rel\_word2vec\_ch1              &     (134.572604613474, 0.4595537826115222)    \\ \hline
rel\_word2vec\_ch2              & (2.8936049625643223, 0.0020496541891822087)\\ \hline
...           & ...            \\ \hline
rel\_word2vec\_ch32              &  (0.9999583161919829, 0.0015334960473239322)       \\ \hline
rel\_baseline\_ch1              &     (-0.771830408650495, 0.008005141647819333)     \\ \hline
rel\_baseline\_ch2              & (-0.008373434318707955, 5.9110606393949324e-05)\\ \hline
...           & ...            \\ \hline
rel\_baseline\_ch32              &  (0.08827132539725344, 0.00013526127447238275)        \\ 
\hline
\end{tabular}
}
\caption{Example of relevance results for the last verb}
\label{Tab:relevance_result_case}
\end{CJK} 
\end{table*}

\begin{table*}
\begin{CJK}{UTF8}{gbsn} 
\centering
\resizebox{1.3\columnwidth}{!}{%
\begin{tabular}{|l|l|}
\hline
\textbf{Regressor}      & \textbf{Example value} \\ \hline
verb                    & 回来 (come back)                     \\ \hline
correct character       & ch4                    \\ \hline
pro-drop                & False                  \\ \hline
salience-glove-unweighted          & 45.761057               \\ \hline
salience-bert-unweighted           & 57.886974               \\ \hline
salience-word2vec-unweighted          & 56.125342               \\ \hline
salience-baseline-unweighted          & 1.087911               \\ \hline
salience-glove-weighted & 1.206085              \\ \hline
salience-bert-weighted  & 1.522071               \\ \hline
salience-word2vec-weighted & 1.427663               \\ \hline
salience-baseline-weighted & 0.979743               \\ \hline
\end{tabular}
}
\caption{Example of salience results for the last verb from three language models and one baseline model with distance-weighted/-unweighted}
\label{Tab:salience-case}
\end{CJK} 
\end{table*}

\end{CJK} 
\end{document}